# Hybrid Embedded Deep Stacked Sparse Autoencoder with w_LPPD SVM Ensemble

Yongming Li*, Yan Lei, Pin Wang, Yuchuan Liu

School of Microelectornics and Communication Engineering, Chongqing University, Chongqing, 400044, China

(Corresponding Author: Yongming Li, yongmingli@cqu.edu.cn )

**Abstract:**

Deep learning is a kind of feature learning method with strong nonliear feature transformation and becomes more and more important in many fields of artificial intelligence. Deep autoencoder is one representative method of the deep learning methods, and can effectively extract abstract the information of datasets. However, it does not consider the complementarity between the deep features and original features during deep feature transformation. Besides, it suffers from small sample problem. In order to solve these problems, a novel deep autoencoder - hybrid feature embedded stacked sparse autoencoder(HESSAE) has been proposed in this paper. HFESAE is capable to learn discriminant deep features with the help of embedding original features to filter weak hidden-layer outputs during training. For the issue that class representation ability of abstract information is limited by small sample problem, a feature fusion strategy has been designed aiming to combining abstract information learned by HFESAE with original feature and obtain hybrid features for feature reduction. The strategy is hybrid feature selection strategy based on $L_1$ regularization followed by an support vector machine(SVM) ensemble model, in which weighted local discriminant preservation projection (w_LPPD), is designed and employed on each base classifier. At the end of this paper, several representative public datasets are used to verify the effectiveness of the proposed algorithm. The experimental results demonstrated that, the proposed feature learning method yields superior performance compared to other existing and state of art feature learning algorithms including some representative deep autoencoder methods.

## 1. Introduction

In the pattern recognition tasks, the strength of the class representation of the features has a decisive effect on the performance of the model. However, redundant information and noise are inevitable for most of the data generated in real life, which will lead to increased complexity and decreased performance of classifier [1]. In addition, it is difficult to obtain the increasingly complex relationships latent in data, which are beneficial to enhance classification performance. Feature learning, a crucial step in pattern recognition tasks, is able to mine key information in data according to specific tasks, while reducing data dimensions. Therefore, it has attracted widely attentions in recent decades [2-4].

Various methods of feature learning have been proposed and they are mainly divided into two categories: feature selection and feature extraction. Feature selection [5] can be seen as a process of searching for an optimal feature subset with strong classification ability according to criteria and obtain high-dimensional characteristics by analyzing low-dimensional data. It is an effective technique to acquire key features from the initial feature space and simplify data analysis. The typical feature selection algorithms include regularization methods -- Least absolute shrinkage and selection operator (Lasso) [6], filter method − Relief [7] and p_value [8], and so on. Feature extraction mainly aims to map original high-dimensional data to a specific low-dimensional space by minimizing information loss. Well-known algorithms contain principle component analysis (PCA) [9], linear discriminant analysis (LDA) [10], locality preserving projections (LPP) [11], etc. PCA projects the original d-dimensional data onto the *l*-dimensional linear subspace ($l < d$), which is spanned by the principal eigenvectors of data's covariance matrix. LDA searches the linear subspace by minimizing the within-class scatter and maximizing inter-class scatter. Both of these two methods are linear dimension-reduction methods. Despite its successful use in computer vision and pattern recognition, the linear model may fail due to most real-world data is nonlinear. LPP, a typical representative algorithm of manifold dimensionality reduction, can optimally preserve the neighboring structure of the dataset and has received extensive attention.

However, large amount of data generated in various fields is not only large in volume but also heterogeneous or complex in nature. Conventional feature learning algorithms are shallow feature learning and rely on empirical knowledge, which cannot effectively mine the inherent non-linear complex relationships between data, thus, there are some limitations for these methods. In recent years, deep feature learning techniques have achieved state-of-the-art performance in various fields, such as image classification [12-13], speech recognition [14-15], machine translation[16-17], etc. Multi-layer neural network has excellent feature learning ability, and the learned features have a more essential description of the data, which is conducive to visualization or classification [18]. Autoencoder (AE) [19], a typical deep learning model, is able to learn data representations by minimizing reconstructed errors between input data and outputs. AEs are easy to stack by taking the outputs of last hidden layer of AE as the inputs of the next AE, which is named stacked autoencoder(SAE). As SAE has advantages of deep neural network and strong expressive ability, SAEs and extensions such as stacked sparse autoencoders(SSAE), stacked denoising autoencoders (SDAE) demonstrate a promising ability to learn meaningful representations[20-22]. SSAEs can find some interesting structures in the input data by introducing sparse constraints, while SDAEs learn useful information by reconstructing input data that contains noise. Although SSAEs and SDAEs have achieved some success in subsequent applications [23-26], effective feature learning of AEs is still a challenging problem. A recent study [27] designed a deep kernelized autoencoder by aligning inner products between codes with respect to a kernel matrix. The proposed autoencoder can learn similarity-preserving embedding of input data. The study [28] proposes an enhanced stacked denoising autoencoder (ESDAE) with manifold regularization for wafer map pattern recognition in manufacturing processes. Specifically, it applies manifold regularization for deep learning to detection and recognition of wafer map defects for further improving performance in describing data distributions. The research [29] proposes to introduce a distance constraint that is added to the SSAE to form a new distance constrained SSAE (DCSSAE) network. They believed that distance constraint can maximize the distinction between the target pixels and other background pixels in the feature space.Yet, there are still some potential drawbacks with these structures. Firstly, it is not always clear what kind of properties of the input data need to be captured by the autoencoders. Not all the features learned by hidden layers are always useful and representative, and sometimes even display a reduced discrimination ability to classification tasks. Secondly, due to a large number of parameters, deep autoencoders are easy to suffer from overfitting with small size sample, which limits the generalization ability of the deep autoencoders to learn effective features.

To tackle the issues mentioned above, a Hybrid Feature Embedded Sparse Stacked Autoencoder(HESSAE) has been proposed here. The primary idea of HESSAE is to evaluate the effectiveness of the outputs in hidden layers and introduce the constraints of original information in the layer-wise training process. To achieve that, original features are embedded into the encoded outputs of each AE to filter out hidden representations with weak discrimination ability. Then the original features are embedded into the output of the current layer (hybrid features) to construct feature representations at higher hidden layers, retaining a certain amount of useful information and being employed to classification task. Sparsity constraint is employed on hidden units to improve the capacity to mine more interesting structure in data. In order to solve the small size sample problem, a feature fusion strategy has been designed. Specifically, feature selection based on $L_1$ regularization is applied on hybrid features output by the HESSAE. In order to further eliminate redundancy of the features and improve the generalization ability of the proposed algorithm, support vector machine (SVM) ensemble model with weighted local discriminant preservation projection (w_LPPD) has been constructed. W_LPPD is a novel manifold learning method proposed by the authors, which considers outliers in the samples and effectively removes some samples far away from the center of the class. The main contributions and innovations of this paper can be stated as follows:

(1) In this paper, hybrid feature embedded unit is introduced into training process for the first time to construct HFESAE, which can obtain the features more complementary with the original features.

(2) In terms of feature fusion, we designed a manifold ensemble learner, which make the deep features learned by HFESAE more efficient for small size samples learning.

(3) Three-step feature learning not only learns deep features with abstract information but also adapts to small samples learning. Meanwhile, it is able to remove low-quality and invalid features, and reduce dimensions, thus improving the accuracy and reliability in subsequent classification tasks.

(4) The hybrid embedded idea can be applied into other deep learning methods except deep stacked autoencoder.

The rest of this paper is organized as follows: section 2 enumerates the related work; section 3 mainly illustrates the proposed algorithm. Experiments and results are presented in section 4 and the final section is discussion and conclusion.

## 2. Related work

In this section, some representative deep autoencoder models and feature fusion methods are reviewed briefly. Deep autoencoder models mainly including typical stacked autoencoder, stacked sparse autoencoder and improved sparse autoencoder. In this paper, the data matrix is denoted as $X=\{x_1, x_2, \cdots, x_N\} \in \mathbb{R}^{N \times n}$ and $x_i = (x_{i1}, x_{i2}, \cdots, x_{in}) \in \mathbb{R}^n$ denotes the $i$-th sample, where $N$ and $n$ are the number of samples and feature dimension respectively. $H = \{h_1, h_2, \cdots, h_N\} \in \mathbb{R}^{N \times d}$ is the output matrix of hidden layer of autoencoder, where $d$ is the dimension of the feature vector in hidden layer.

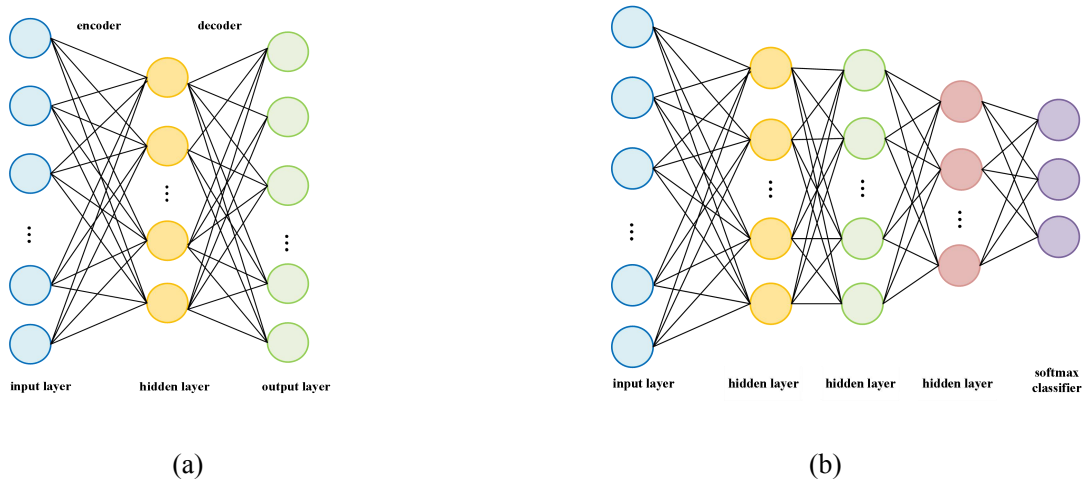

Fig.1. (a) autoencoder(AE); (b) stacked autoencoder(SAE).

### 2.1 Representative Stacked Autoencoder models
*A. Stacked Autoencoder(SAE)*

Autoencoder (AE) [19] is one type of artificial neural networks structurally defined by three layers: input layer, hidden layer and output layer, which compose an encoder and a decoder. In the AE, the encoder transforms date matrix $X$ into a hidden representation $H$ with a tunable number of neural units $d$, and fallowed by a nonlinear activation.

*B. Stacked Sparse Autoencoder(SSAE)*

Autoencoder is originally proposed based on the idea of dimensionality reduction, which will lose the ability to automatically learn features when there are more hidden layer nodes than the input nodes. Sparse autoencoder(SAE) [31] is obtained by adding some sparse constraints on the traditional autoencoder, that is suppress most of the output of hidden layer neurons.

*C. Locality-Constrained Sparse Autoencoder(LSAE)*

Luo et al. [32] proposed a locality-constrained sparse auto-encoder (LSAE) for image classification. They believe that the locality is more essential than sparsity for classification tasks. Therefore, the concept of locality is introduced into the sparse autoencoder, which ennables the autoencoder to encode similar inputs using similar features.

*D. Weight-Clustering Sparse Autoencoder(WCSAE)*

Fan et al. [33] proposed a weight-clustering sparse autoencoders(WCSAE) combined object-oriented classification with difference images. They believed that superfluous features are extracted in high-level feature extraction when similar weights exist in the weight matrix, which cannot enhance the performance to represent data. Therefore, similar weights in the hidden layer of WCSAE are clustered layer-wise.

*E. Stacked Pruning Sparse Autoencoder(SPSAE)*

Haiping Zhu et al. [45] proposed a new stacked pruning sparse autoencoder (SPSAE) model. The main idea of this model is to uses the superior features extracted in all the previous layers to participate in the subsequent layers, thus to create new channel between the front layers and the back layers. In addition, a pruning operation is added into the model so as to prohibit non-superior units from participating in all the subsequent layers. Specifically, the SPSAE is constructed by SAE model, whose training objection is followed by Eq.9. Theses SAEs are constructed as a fully connected network rather than simply connected in series, which is, directly connecting the layers with feature mapping in the network. In this way, the input of the each layer comes from the output of all the previous layers and the feature information of previous layers can be shared to subsequent layers. To solve the problem of the increased number of SAE units and training time, the input data of the units with large reconstruction errors in pre-training are prohibited from participating in the subsequent training process, which is called pruning operation.

2.2 Deep-Shallow Feature Fusion

In recognition tasks, combining relevant information of multiple features is divided to decision level fusion and feature level fusion. Majtner et al. [34] proposed a decision level fusion algorithm for skin lesion classification. They combined two SVM classifier. One used hand-crafted features, and another employed feature derived from CNN. Both of the two classifier predicted the class or class score, then the label with the highest absolute score value would be chosen as result. To classify the thyroid nodules, Liu et al. [35] also fused deep features learned by CNN and three kinds of conventional features in decision level, but the classification result was determined by positive-sample-first majority voting strategy.

In terms of feature level fusion, Suk et al. [36] extracted abstract information in raw handcrafted imaging features by deep autoencoder as deep features, which were concatenated with raw feature vectors. Sparse learning was employed on cascaded feature to select optimal subset, which is send to clssifier for AD/MCI diagnosis. Mei et al. [37] proposed an unsupervised-learningbased feature-level fusion approach for mura defect recognition. Features extracted by the unsupervised-learning-based model and the handcrafted descriptors are simply concatenated to form the representations of defect images. Kan [38] proposed a general fusion unit to fuse handcrafted feature information and CNN representations. Specifically, both of them were transformed into a relatively consistent space by fully connected converter, respectively, which can suppress useless information. Then, merge two type features by a supervised feature embedding layer.

## 3. The proposed method

In this section, the proposed algorithm is introduced here. This algorithm mainly includes three parts. First, a proposed hybrid feature embedded stacked sparse autoencoder(HESSAE) is proposed for learning the feautres with high complementary with original features. Seconded, $L_1$ regularization learning is designed and applied on the hybrid features (output of HESSAE and original features) to select optimal feature subset. Finally, an ensemble model is established based on w_LPPD and SVM. The three-step feature leaning method is able to extract abstract information as well as retain initial information and remove redundancy among features.

### 3.1 Hybrid-Feature embedded sparse stacked autoencoder (HESSAE)

The proposed hybrid-feature embedded stacked sparse autoencoder(HESSAE) model is given in Fig. 2.

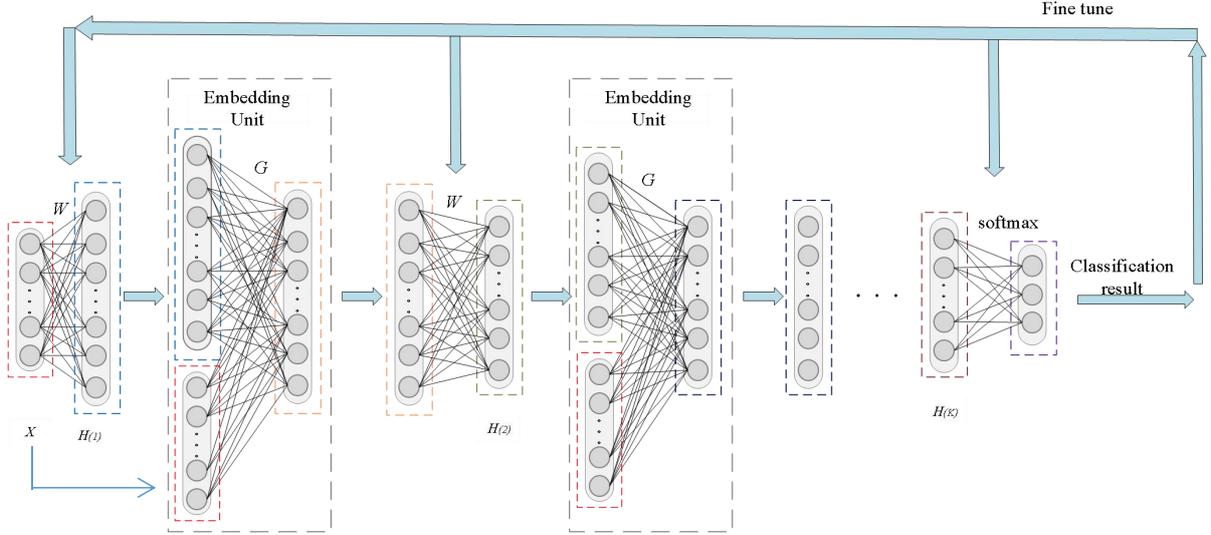

Fig.2. Hybrid feature embedded stacked sparse autoencoder (HESSAE)

The embedding unit between two encoders is an important component for the proposed model. Assuming HESSAE is constructed by $K$ autoencoders, and $H^{(k)} = \{h_1^{(k)}, h_2^{(k)}, \cdots, h_N^{(k)}\} \in \mathbb{R}^{N \times d^{(k)}}$ denotes the output matrix of $k$-th hidden layer. The first autoencoder in HESSAE is traditional autoencoder training The coded feature vectors in the rest layers of HESSAE can be defined as:

$$h^{(k)} = f(W_{k1}^T L(x \oplus h^{(k-1)}) + b_{k1}), \quad 1 < k \leq K \quad (1)$$

Where $W_{k1}$ and $b_{k1}$ are weight matrix and bias vector of encoder in $k$-th autoencoder respectively. In this paper, logistic sigmoid function $f(x) = \dfrac{1}{1+\exp(-x)}$ is used as the activation function.

In Eq. (1), $\oplus$ represents concatenating learned feature vector and original feature vector, and $L$ denotes an embedding units, which can be formulated by:

$$L(x \oplus h^{(k-1)}) = G^T(x \oplus h^{(k-1)}) \quad (2)$$

Where $G \in \mathbb{R}^{(n+d^{(k-1)}) \times d^{(k-1)}}$ is the corresponding transformation matrix consisted by zero and one. The embedding units aims to introduce certain initial information constraint during layer-wise pre-training process, meanwhile, filter out some useless features. Considering the divergence of feature can explain their class discrimination ability to some extends, the objective function of embedding units can be defined as:

$$\max_{G} tr(G^T(X \oplus H^{(k-1)})(X \oplus H^{(k-1)})^T G)$$
$$s.t \sum G_{ij} = d^{(k-1)} \quad (3)$$

Where $d^{(k)}$ is the number of hidden units in $k$th layer. Calculating covariance matrix $D$ of $X \oplus H^{(k-1)}$ and sorting the diagonal elements in decreasing order. Then the first $k-1$ values are taken to form a vector $s \in \mathbb{R}^{(k-1)}$. The elements of $G$ can be determined by:

$$G_{ij} = \begin{cases} 1, & \text{if } s_j = D_{ii} \\ 0, & \text{otherwise} \end{cases} \quad (4)$$

Where $D_{ii}$ is the $i$th diagonal element of covariance matrix $D$, $s_j$ is the $j$-th element of $s$, and $i \in [1, n+d^{(k-1)}], j \in [1, d^{(k-1)}]$. With Eq. (1-4), the hidden outputs of each autoencoder can be obtained. The decoder function of $k$th autoencoder can be rewritten as:

$$L'(x \oplus h^{(k-1)}) = f(W_{k2}^T h^{(k)} + b_{k2}), \quad 1 < k \leq K \quad (5)$$

Where $L'$ denotes the reconstruction for the output of embedding unit. $W_{k2}$ and $b_{k2}$ are weight matrix and bias vector of $k$-th decoder. In the training process, sparse criterion is applied to hidden layers of HESSAE to discover interesting structures in input data. In this paper, Kullback-Leibler (KL) divergence as a tractable unsupervised objective is introduced in order to sparse representation. It is the relative entropy to measure the difference between two Bernoulli random variables: the average activation $\hat{\rho}_j$ of the $j$-th hidden unit and the target average activation $\rho$. KL divergence is formulated as follows:

$$\sum_{j=1}^{d} KL(\rho \| \hat{\rho}_j) = \sum_{j=1}^{d} \rho \log(\frac{\rho}{\hat{\rho}_j}) + (1-\rho) \log(\frac{1-\rho}{1-\hat{\rho}_j})$$
$$\hat{\rho}_j = \frac{1}{N} \sum_{i=1}^{N} f^j(x^{(i)}) \quad (6)$$

Where $f^j(x^{(i)})$ is the activation value of the $i$-th input vector to the $j$-th unit of hidden layer. The value increases monotonously along with the increasing different between $\rho$ and $\hat{\rho}_j$, and when $\hat{\rho}_j = \rho$, $KL(\rho \| \hat{\rho}_j) = 0$. Therefore, by setting a small sparse parameter $\rho$, most of the average outputs of hidden units are zeros and, thus, sparse representation can be achieved. By introducing embedding units and sparse constraint, the training object function of $k$th ($k > 1$) layer in HESSAE can be written as follows:

$$\arg\min_{\theta} \frac{1}{N} \sum_{i=1}^{N} \| L(x_i \oplus h_i^{(k-1)}) - L'(x_i \oplus h_i^{(k-1)}) \|^2 + \lambda(\| W_{k1} \|_2 + \| W_{k2} \|_2) + \beta(\sum_{j=1}^{d^{(k)}} \rho \log(\frac{\rho}{\hat{\rho}_j}) + (1-\rho) \log(\frac{1-\rho}{1-\hat{\rho}_j})) \quad (7)$$

Where $\lambda$ and $\beta$ denote Penalty parameters of regularization term and sparsity constraint respectively. Training process with eq. (18) is called pre-training, after which the hidden layer of pre-trained autoencoders are cascaded one by one to form a stacked autoencoder, and its initial parameters are determined by pre-training. Focusing on the ultimate goal is to obtain features with better category representation ability, so we further optimize the whole network in a supervised manner. In order to achieve that, softmax layer are taken as classification layer connected on the top of the HESSAE. The fine-tune process of the stacked network is based on back-propagation with gradient decent. Because of the characteristics of pre-tune, the fine-tune flowing it can reduce the risk of falling into local optimum. The proposed HESSAE algorithm is summarized in Algorithm 1.

The learned nonlinear transformation by HESSAE can be regard as a effective feature learning, which not only makes use of the characteristics that deep network can learn the potential relationships among data, but also improves the robustness of deep features by introducing initial information constraint into the deep network. After the training of the whole network, for every input original feature vector $x_i = (x_{i1}, x_{i2}, \cdots, x_{in})$, a new feature vector can be obtained in each hidden layer and different layer represents different level information. Generally, the higher the layer in the network, the more complicated or abstract patterns inherent in the input data there is. Therefore, we take the outputs of the last hidden layer as learned deep feature, recorded as $x_i' = (x_{i1}', x_{i2}', ..., x_{iq}')$. Then we construct an augmented feature vector $\hat{x}_i$ by concatenating $x_i$ and $x_i'$:

$$\hat{x}_i = (x_{i1}, x_{i2}, \cdots, x_{id}, x_{i1}', x_{i2}', ..., x_{iq}') \in \mathbb{R}^{(n+q)} \qquad (8)$$

| **Algorithm 1: HESSAE** |
|---|
| **Input**: training data $X$ |
| Step1: Setting parameters, including: $\lambda$, $\beta$, $\rho$, and the number of hidden units in each AE, number of iteration. |
| Step2: Pre-training: <br> 1) Training the first layer of HESSAE and extracting the output of encoder as $H^{(1)}$. <br> 2) for k =2:K <br>        a) Embedding original feature into deep feature $H^{(k-1)}$ by Eq(2); transformation matrix G is determined by Eq. (3-4). <br>        b) Training the *k*th AE with object function in Eq. (7), then extracting the output $H^{(k)}$ of hidden layer, which will be the input of next AE. <br>      end |
| Step3: Stacking the hidden layer and adding an softmax layer on the top, of which the input is $H^{(k)}$ |
| Step4: Fine-tune the whole network based on back-propagation with gradient decent, and the training goal is to minimize classification loss. |
| Step5: Extracting the output $H^{(k)}$ of final hidden layer in the fine-tuned HESSAE as learned deep feature. |
| **Output**: Deep feature |

## 3.2 Hybrid feature selection strategy based on L$_1$ regularization

The hybrid feature vector obtained according to Eq. (8) has richer category information. However, simple combination will lead to high dimension problem, which means possible dimension disaster. On the other hand, we believe that these features are not independent, and there is some redundant information among two types of feature. Therefore, it is necessary to process the candidate feature vector effectively to extract the most useful information.

Specifically, $L_1$ regularization uses a penalty term to control the sum of the absolute values of the parameters to be small, giving a sparse feature vector. For the new dataset $\hat{X} = \{(\hat{x}_i, y_i)\}_{i=1}^{N}$, where $\hat{x}_i \in \mathbb{R}^{n+q}$ denotes $i$th sample with hybrid features and $y_i$ is corresponding label. Considering the simplest regression model with the squared error as loss function, the optimization objective function can be defined as:

$$\arg\min_{w} \sum_{i=1}^{N}(y_i - \sum_{p=1}^{n+q} w_p \hat{x}_{ip})^2 + \kappa \sum_{p=1}^{n+q} |w_p| \quad (9)$$

Where $L_1$-regularization is introduced to mitigate the problem of overfitting. $w_p$ is the regression coefficients of $p$th feature. $\kappa$ denotes a sparsity control parameter, and the larger it is, the more sparse the model is. Proximal Gradient Descent is used to solve Eq. (9), and the iteration of each step should be:

$$w^{(k+1)} = \arg\min_{w} \frac{M}{2} \| w - (w^{(k)} - \frac{1}{M} \frac{\partial(\sum_{i=1}^{N}(y_i - (w^{(k)})^T \hat{x}_i)^2)}{\partial w^{(k)}}) \|_2^2 + \kappa \| w \|_1 \quad (10)$$

Where $w = (w_1, w_2, \cdots w_{n+q})$, $M$ is a constant great than zero.

Assuming $u = w^{(k)} - \frac{1}{M} \frac{\partial(\sum_{i=1}^{N}(y_i - (w^{(k)})^T \hat{x}_i)^2)}{\partial w^{(k)}}$, the closed-form solution of Eq. (10) can be calculated by:

$$w_p^{(k+1)} = \begin{cases} u_p - \kappa/M, & \kappa/M < u_p; \\ 0, & |u_p| \leq \kappa/M; \\ u_p + \kappa/M, & u_p < -\kappa/M; \end{cases} \quad (11)$$

In Eq. (11), $w_p^{(k+1)}$ and $u_p$ are the $p$-th component of $w^{(k+1)}$ and $u$ respectively. The result of solving the $L_1$-normalization expresses that only features corresponding to the non-zero component of $w_p$ can be choose to the final feature subset.

### 3.3 w_LPPD SVM ensemble model
### 3.3.1 Weighted locality preserving discriminant projection（w-LPPD）

Weighted local preserving discriminant projection is a novel effective feature reduction method, which considers outliers in the samples, removing some samples far away from the center of the class. Firstly, it introduces random subspace sampling; Secondly, locality preserving discriminant projection is established based on the proposed objective function; finally, integrating multi-space mapping matrices to construct the ultimate mapping matrix. Assuming $k_{mc}$ denotes the number of samples sampling for $c$-th, the number of total samples after sampling is $k_m = \sum_{c=1}^{C} k_{mc}$. The local between-class scatter matrix $S_{LB}$ with the $k_m$ nearest neighbors of the center $\mu_{lb}$, and the local within-class scatter matrix $S_{LW}$ with the $k_{mc}$ nearest neighbors of the class center $\mu_{lwc}$ can be defined as follows:

$$S_{LB} = \sum_{c=1}^{C} N_{lc}(\mu_{lbc} - \mu_{lb})(\mu_{lbc} - \mu_{lb}) \quad (12)$$

$$S_{LW} = \sum_{c=1}^{C} \sum_{i=1}^{k_{mc}} (x_i^{(c)} - \mu_{lwc})(x_i^{(c)} - \mu_{lwc})^T \quad (13)$$

Where local numbers $k_m = \lfloor r_b \cdot N \rfloor$ and $k_{mc} = \lfloor r_w \cdot N_c \rfloor$, $r_b$ and $r_w$ are sampling ratio coefficients, $N$ and $N_c$ are the number of total sample and c-th sample respectively. $\mu_{lb} = \frac{1}{k_m} \sum_{i=1, x \in N_{k_m}(m)}^{k_m} x_i$ is the center of local part for $S_{LB}$ computation, $\mu_{lbc} = \frac{1}{N_{lc}} \sum_{i=1, x \in N_{k_m}(m)}^{N_{lc}} x_i^{(c)}$ is the center of the c-th local part for $S_{LB}$ computation, $N_{lc}$ is the number of the c-th class in local part, and $\mu_{lwc} = \frac{1}{k_{mc}} \sum_{i=1, x^{(c)} \in N_{k_{mc}}(m_c)}^{N_{lc}} x_i^{(c)}$ is the center of the c-th local class for $S_{LW}$ computation.

Furthermore, the locality preservation regularization term is shown as follows:

$$\begin{aligned} \min_W \sum_{i=1}^{N} \sum_{j=1}^{N} A_{ij} \|W^T x_i - W^T x_j\|^2 \\ = 2Tr(W^T \sum_{i=1}^{N} \sum_{j=1}^{N} A_{ij}(x_i x_i^T - x_i x_j^T) W) \\ = Tr(W^T XLX^T W) \end{aligned} \quad (14)$$

Where $L$ is the Laplacian matrix, $D_{ii} = \sum_j A_{ij}$ is the diagonal matrix, and $A$ is the affinity matrix, which can be calculated in following manners:

$$A_{ij} = \begin{cases} 1, & \text{if } x_i \in N_k(x_j) \| x_j \in N_k(x_i) \\ 0, & \text{others} \end{cases} \quad (15)$$

With Eq. (12-14) and the proposed w-LPPD can be formulated as:

$$\begin{aligned} \min_W Tr(W^T S_{LW} W) \\ s.t. \ Tr(W^T S_{LB} W) - \gamma Tr(W^T XLX^T W) = \alpha I \end{aligned} \quad (16)$$

Where $\gamma$ represent the regularization coefficient, $\alpha$ is constant. It can be seen from the objective function that the w-LPPD aims to minimize the trace of local within-class scatter matrix and maximize the between-class scatter matrix while preserving the locality of the sample.

By introducing Lagrange multiplier $\lambda$, the objection function Eq. (16) finally can be written as:

$$L(W,\lambda) = Tr(W^T S_{LW} W) - \lambda(W^T S_{LB} W - \gamma W^T XLX^T W - \alpha I) \qquad (17)$$

Taking the derivation of $W$ and obtain optimal solutions.

$$\frac{\partial L(W,\lambda)}{\partial W} = 0$$
$$\Rightarrow \frac{\partial \{Tr(W^T S_{LW} W) - \lambda(W^T S_{LB} W - \gamma W^T XLX^T W - \alpha I)\}}{\partial W} = 0 \qquad (18)$$
$$\Rightarrow (S_{LB} - \gamma XLX^T)^{-1} S_{LW} W = \lambda W$$

Apparently, through Eq. (18), the projection matrix $W$ can be easily obtained by generalized eigenvalue decomposition. The vector $W_k = (w_1, w_2, ..., w_k)$ is comprised of the top $k$ eigenvectors of $W$. Then, the original data can be projected into a low dimension space spanned by the columns of $W_k$ to achieve dimensionality reduction. As mentioned before, we exploit LPPD on random subspace, so we can get $P$ projection matrixes: $W_k^1, W_k^2, \cdots, W_k^p$. The final mapping matrix $W_k^F$ is obtained by weighting $W_k^1, W_k^2, \cdots, W_k^p$. Its mathematical expression as follows:

$$W_k^F = \alpha_1 W_k^1 + \alpha_2 W_k^2 + \cdots + \alpha_p W_k^p = \sum_{i=1}^{p} \alpha_i W_k^p \qquad (19)$$

Where $\alpha_i$ is the weight coefficient, it can be determined by grid search. Note that $\sum_{i=1}^{p} \alpha_i = 1$.

Through w_LPPD, we can further map the deep-shallow feature subset selected by $L_1$ regularization to another low-dimension feature space, where the distance of samples from different class will be farther, but the distance of samples from same class will be closer. According to that, the feature obtained by this way own more effective class representation and discriminant ability

### 3.3.2 Ensemble learning based on w_LPPD and SVM

Assuming that sampling ration of samples and features are $\delta_1$ and $\delta_2$ respectively, and sampling $K$ times to form $K$ subsets based on bagging strategy. Then w_LPPD is applied on each subsets. The ultimate $K$ training subsets obtained by w_LPPD will respectively be fed into $K$ base classifiers for training. In this paper, support vector machine(SVM) is used as base classifier. The classification result of validation samples will be decided by weighting voting mechanism. The weight of each classifier can be calculated according to the following formula:

$$w_k = \frac{\sum_{i=1}^{N_{train}} \phi(C_{ik}, y_i)}{N_{train}} \qquad (20)$$

Where $\phi(C_{ik}, y_i) = \begin{cases} 1, & if\ C_{ik} = y_i \\ 0, & others \end{cases}$, $N_{train}$ means the number of training set. Assuming the class number of dataset is $C$, for $i$-th sample $x_i$ with label $y_i$, $C_{ik}$ is the forecast result of k-th classifier. The probability of sample $x_i$ belongs to $c$-th class can be expressed as:

$$P(x_i \in x^c)|_{c=1}^{C} = \sum_{k=1}^{K} w_{kc} \phi(C_{ik}, c) \tag{21}$$

Then the ultimate class label predicted by the ensemble model can be decided by following formula:

$$y_i' = \max_c \{P(x_i \in x^1), P(x_i \in x^2), \cdots, P(x_i \in x^C)\} \tag{22}$$

## 4. Experimental result and analysis
### 4.1 Data and Experimental conditions

In our study, groups of experiments have been performed on several datasets with different sample size and feature dimension to validate the performance of the proposed method. Brief information about dataset is shown in Table I.

All experiments were carried out in an experimental environment: the experimental operating system is Windows 10, 64-bit operating system, and the memory size was 128 GB. The programming tool is MATLAB, vision R2018b.

For description, the number of base autoencoder in the proposed HESSAE is set as three. The number of each hidden units is determined by taking into account of the dimensionality of the original feature vector for different datasets to choose range and using grid search to find the best structure. It is worth noting that different structure for different dataset reflects the necessity of considering different high-level non-linear relations inherent in low-level feature for different classification tasks. The tunable parameters in the proposed HESSAE model mainly contains regularization parameter $\lambda$, $\beta$ and sparsity parameter $\rho$. The optimal structure and parameters of the HESSAE models in our experiments are presented in table II.

Table I. Basic information of datasets

| Database name | Instances | Attributes | Class | Relevant papers |
|---|---|---|---|---|
| AD | 90 | 32 | 3 | Reference [39] |
| Statlog_Landsat_Satellite (Statlog) | 6435 | 36 | 6 | Reference [41] |
| Pen-Based Recognition of Handwritten Digits (Pen-Digits) | 10992 | 16 | 10 | Reference [42] |
| Urban land cover (Urban) | 675 | 148 | 9 | Reference [44] |

Table II. The HESSAE structure and parameters for seven datasets

| datasets | Hidden units | $\lambda$ | $\beta$ | $\rho$ | epochs |
|---|---|---|---|---|---|
| AD | 100 - 50 - 25 | $10^{-5}$ | 5 | 0.02 | 500 |

| | | | | | | |
|---|---|---|---|---|---|---|
| Urban | 600 - 300 - 80 | $10^{-3}$ | 2 | 0.07 | 600 |
| Pen-Digits | 80 - 30 - 10 | $10^{-4}$ | 4 | 0.05 | 1000 |
| Statlog | 120 - 60 - 20 | $10^{-3}$ | 5 | 0.05 | 1000 |

For w_LPPD SVM ensemble, local ratio coefficients in w_LPPD are set as $r_b = 0.9$, $r_w = 0.9$, the sampling ration $\delta_1 = 0.7$, $\delta_2 = 0.5$, and the number of base classifier is set as five. In experiment, hold-out cross validation method was used. That means for all datasets, the labeled samples were split into two subsets, one accounted for one-third of all samples as test data, and the rest as train data. In order to eliminate the influence of accidental factors, each experiment was repeat five times to calculate the average accuracy and standard deviation as ultimate performance.

### 4.2 Verification of the proposed method

In order to reflect the efficiency of the proposed algorithm, our methods are compared with representative feature learning methods, including feature selection algorithms: Lasso, relief and p_value; feature extraction algorithms: PCA, LDA, and LPP. To ensure a fair comparison, in each experiment, the training samples and test samples for each method are identical, and grid-search is used to find the best parameter for these methods. Taking into account of the base classifier in our methods is SVM, so we adopted SVM as classifier to evaluate aforementioned approaches. The average accuracy and variance are listed in table Ⅲ.

Table Ⅲ illustrated that the proposed method have a superior performance compared to others. For six datasets, our methods can significantly improve classification accuracy. For example, the classification accuracies on AD and LSVT are enhanced by more than 12% compared with no processing, and they are also outperform traditional feature selection and feature extraction methods. In addition, as what can be seen, the standard deviations under our method are significantly smaller, which means that our algorithm is more stable. The possible reason is that the proposed method is capable to discover potential patterns among the data and effectively fuse the low-level and high-level features followed by dimension-reduction ensemble model, thus improving classification accuracy obviously. We also can see the proposed method have better generalization capability.

Table Ⅲ. Classification accuracy (mean ± variance) of different algorithms (%)

| | OF | PCA | LDA | LPP | Rlief | Lasso | P_value | Proposed method |
|---|---|---|---|---|---|---|---|---|
| AD | 54 ±9.54 | 60 ±8.49 | 60 ±7.07 | 58.67 ±6.91 | 56.66 ±8.49 | 55.33 ±8.02 | 48.67 ±7.67 | **67.33 ±2.49** |
| Urban | 79.99 ±1.04 | 81.51 ±1.02 | 82.57 ±0.73 | 81.96 ±0.81 | 81.15 ±0.87 | 79.64 ±0.96 | 79.82 ±1.79 | **83.20 ±1.01** |
| Pen-Digits | 97.52 ±0.22 | 97.49 ±0.23 | 97.37 ±0.35 | 97.91 ±1.14 | 96.94 ±0.26 | 97.52 ±0.22 | 93.07 ±2.21 | **98.00 ±0.12** |
| Statlog | 86.64 ±0.41 | 87.43 ±0.57 | 87.23 ±0.42 | 87.54 ±0.59 | 86.59 ±0.22 | 86.61 ±0.33 | 86.63 ±0.29 | **87.28 ±0.12** |

**Notes** OF: Original features

In order to verify the feature extraction capability of the proposed hybrid feature embedded sparse stacked

autoencoder(HESSAE), stacked autoencoder(SAE), stacked sparse autonencoder(SSAE) are compared with the proposed network. The proposed autoencoder without sparse constraints named HESAE, which is used as a comparison to verify the necessity of sparse constraints for HESSAE. All the networks are constructed with three hidden layers and a softmax layer. The regularization parameters and sparse parameter are set to the same values. The classification accuracy of the deep features learned by the above structures are recorded in Table Ⅳ. As shown in table Ⅳ, the proposed method structure achieved the highest accuracy in most cases, which proves the idea of introducing initial information into and filtering out some less discriminative features in pre-processing is feasible. In table Ⅳ, we also can observe that sparse constraints can significantly improve the robustness of deep autoencoders, no matter the conventional one or the proposed one.

Table Ⅳ Classification accuracy (mean ± variance) of different deep autoencoder classifiers(%)

|  | SAE[20] | SSAE[31] | Proposed HESSAE | Proposed HESAE |
|---|---|---|---|---|
| AD | 50.67±7.95 | 56.67±5.27 | **59.33±3.65** | 56.00±4.94 |
| Urban | 74.48±3.33 | **79.73±0.67** | 71.91±2.63 | 68.00±3.09 |
| Pen-Digits | 89.64±1.44 | 93.80±0.51 | **96.04±0.47** | 94.07±0.53 |
| Statlog | 83.67±0.36 | 84.85±0.84 | 86.36±0.59 | **86.39±0.43** |

Furthermore, the major parts of the proposed method are verified. To verify the high-level feature learned by HESSAE can been regarded as the latent representations with discriminant information hidden in the data, we designed experiments that only use high-level feature. To validate the combination of low-level feature and high-level feature will result in high dimensionality and high redundancy, simultaneously show our methods can mitigate these problems gradually, we set up experiments that use merged features without process, merged feature with $L_1$ regularization, and merged feature with the proposed fusion mechanism, respectively. The results are listed in tableⅤ.

As we can see, deep features do have a certain class representation ability, and sometimes they are similar to or even exceeded the original features. In deep learning, a large number of training samples is required to prevent overfitting, so datasets with relatively larger size benefit from HESSAE, such as Pen-digits with 10992 samples and Statlog with 6435 samples. Cascading raw data directly with deep features have shown lower accuracy in almost every datasets, which we believe is caused by high dimensionality information redundancy. The results demonstrated that $L_1$ regularization and w-LPPD SVM ensemble employed on hybrid feature are able to remove redundancy effectively and retain valid classification information. Meanwhile, the designed fusion method is an effective fusion strategy that can improve the accuracy and stability of the classification. The results show that the major parts are effective.

Table Ⅴ. Classification accuracy (mean ± variance) of the major parts in the proposed method (%)

|  | OF | DF | HF | HF&$L_1$ | HF&$L_1$& Ensemble |
|---|---|---|---|---|---|
| AD | 54.00 ±9.54 | 57.33 ±5.47 | 52.00 ±6.05 | 53.33 ±6.33 | **67.33 ±2.49** |
| Urban | 79.99 ±1.04 | 71.64 ±1.45 | 79.73 ±1.31 | 80.27 ±1.07 | **83.20 ±1.01** |
| Pen-Digits | 97.52 ± 0.22 | 96.04 ± 0.67 | 97.51 ± 0.21 | 97.76 ± 0.18 | **98.00 ± 0.12** |
| Statlog | 86.64 ± 0.41 | 86.45 ± 0.74 | 86.60 ± 0.43 | 86.68 ± 0.39 | **87.28 ± 0.12** |

**Notes:** DF: Deep features learned by HESSAE; HF: Hybrid feature (Cascade of original features and deep features); HF&$L_1$: Hybrid feature selection with $L_1$ regularization; HF&$L_1$&Ensemble : HF&$L_1$ followed by w_LPPD SVM ensemble.

## 5. Discussion and Conclusion

Deep stacked autoencoder is a kind of important deep learning method. Although it is easy to understand and realize, and has wide application, existing deep stacked autoencoder methods did not consider to fuse the original features within hidden layers and during training. One of the negative results is the deep features are not complementary with the original features sufficiently, thereby affecting the quality of subsequent feature fusion. To tackle this issue, a novel improved deep stacked autoencoder ensemble has been proposed in this paper. Deep features extracted by an improved hybrid feature embedded stacked autoencoder are fused with original features to mitigate the small-sample problem, then feature selection based on $L_1$ regularization and w_LPPD SVM ensemble model have been designed for solving high dimensionality problem and improving reliability. Experiments were conducted on various representative datasets with different sample size and feature dimension. The experimental results demonstrated the proposed method outperforms other well-known methods in terms of classification accuracy and stability.

The value of this work can be summarized as follows:(1) The proposed HESSAE can obtain better initial parameters by embedding initial information and discarding some weak features during pre-training, thus learning more effective deep representations and more complementary features with original features. (2) W_LPPD-SVM ensemle is designed to deal with feature subset with small size samples. (3) Three-steps feature learning is designed and is able to extract abstract information as well as retain useful initial information. The advantages above contribute to higher accuracy and stability of the proposed algorithm. 4) the hybrid embedded idea can be applied into other deep learning methods except deep stacked autoencoder.

There are still some shortcoming of the proposed methods, the future work is to optimize the structure or training standards of deep encoders, improving the quality of learned deep features. Besides, other deep learning methods can be considered for further verification.

## Acknowledgments

We are grateful for the support of the National Natural Science Foundation of China NSFC (No. 61771080); the Fundamental Research Funds for the Central Universities (2019CDQYTX019, 2019CDCGTX306), the Basic and Advanced Research Project in Chongqing (cstc2018jcyjAX0779); and the Southwest Hospital Science and Technology Innovation Program (SWH2016LHYS-11).

## Author contributions

YL conceived of the study, participated in its design and coordination and helped draft the manuscript. YL and PW participated in the measurements of all subjects and drafted the complete manuscript. YL managed the trials and assisted with writing the discussion in the manuscript.

All authors read and approved the final manuscript.

## Conflict of Interest

The authors declare that they have no conflicts of interest related to this work.

## Ethics Approval and Consent to Participate

Not applicable

**Consent for publication**

Not applicable